\documentclass[]{bytedance_seed}

\usepackage[toc,page,header]{appendix}
\usepackage{wrapfig}

\usepackage{listings} 

\lstdefinestyle{rawwrap}{
frame=single,              
    language={[plain]tex}, 
    basicstyle=\ttfamily,  
    columns=fullflexible,  
    keepspaces=true,       
    breaklines=true,       
    breakautoindent=false, 
    postbreak={}, 
}

\usepackage{minitoc}

\title{\fontsize{16pt}{20pt}\selectfont Reverse-Engineered Reasoning for Open-Ended Generation}

\author[1,2,4]{$^\dagger$Haozhe Wang$^+$}
\author[1,3]{Haoran Que$^+$}
\author[6]{Qixin Xu}
\author[4,5]{Minghao Liu}
\author[4]{Wangchunshu Zhou}
\author[1]{Jiazhan Feng}
\author[1]{Wanjun Zhong}
\author[3]{Wei Ye}
\author[3]{Tong Yang}
\author[1]{Wenhao Huang}
\author[1,4]{$^\dagger$Ge Zhang}
\author[2]{$^\dagger$Fangzhen Lin}

\affiliation[1]{ByteDance Seed}
\affiliation[2]{Hong Kong University of Science and Technology}
\affiliation[3]{Peking University}
\affiliation[4]{M-A-P}
\affiliation[5]{2077AI}
\affiliation[6]{Tsinghua University}

\contribution[+]{Work done at ByteDance Seed}
\contribution[\dagger]{Corresponding authors}

\abstract{
While the "deep reasoning" paradigm has spurred significant advances in verifiable domains like mathematics, its application to open-ended, creative generation remains a critical challenge. The two dominant methods for instilling reasoning—reinforcement learning (RL) and instruction distillation -- falter in this area; RL struggles with the absence of clear reward signals and high-quality reward models, while distillation is prohibitively expensive and capped by the teacher model's capabilities. To overcome these limitations, we introduce REverse-Engineered Reasoning (REER), a new paradigm that fundamentally shifts the approach. Instead of building a reasoning process "forwards" through trial-and-error or imitation, REER works "backwards" from known good solutions to computationally discover the latent, step-by-step deep reasoning process that could have produced them. Using this scalable, gradient-free approach, we curate and open-source DeepWriting-20K, a large-scale dataset of 20,000 deep reasoning trajectories for open-ended tasks. Our model, DeepWriter-8B, trained on this data, not only surpasses strong open-source baselines but also achieves performance competitive with, and at times superior to, leading proprietary models like GPT-4o and Claude 3.5.
}

\date{\today}
\correspondence{Ge Zhang at \email{zhangge.eli@bytedance.com}, Haozhe Wang at \email{jasper.whz@outlook.com}}

\checkdata[Project Page]{\url{https://m-a-p.ai/REER_DeepWriter}}

\begin{document}
\maketitle


\definecolor{rliableolive}{HTML}{BBCC33}
\definecolor{rliableblue}{HTML}{77AADD}
\definecolor{rliablered}{HTML}{EE8866}
\definecolor{SDEblue}{RGB}{28 58 88}
\definecolor{cc1}{rgb}{1.0, 0.44, 0.37}
\definecolor{cc2}{rgb}{0.0, 0.2, 0.6}
\definecolor{cc3}{RGB}{255, 191, 0}
\definecolor{cc4}{RGB}{0, 128, 128}

\section{Introduction}
The paradigm of "deep reasoning" is catalyzing a shift in large language model (LLM) reasoning, moving beyond rapid, surface-level inference to leverage increased computational investment at test time~\cite{deepseek, oai, qwq, s1, deepthink}. This approach unlocks sophisticated capabilities like multi-step planning and complex problem-solving, yielding remarkable performance gains in verifiable domains such as mathematics and programming. The success in these areas has been largely propelled by Reinforcement Learning (RL), where clear reward signals for correct outcomes effectively guide a model’s search through vast solution spaces.

However, the reliance on verifiability presents a formidable barrier when applying deep reasoning to open-ended, creative domains~\cite{writingzero, rlhf}. Creative writing, a quintessential example of a non-verifiable task, lacks a singular, objective ground truth. Instead, its quality is judged on subjective criteria like originality, emotional resonance, and narrative coherence~\cite{writingbench}. This disconnect raises a critical and largely unexplored research question:

\begin{tcolorbox}[colback=rliableolive!10!white,colframe=black,boxrule=1pt,boxsep=2pt,top=3pt,bottom=3pt,left=2pt,right=2pt]
\begin{center}
\textbf{How can we instill deep reasoning for open-ended generation in the absence of task verifiability?}
\end{center}
\end{tcolorbox}

Bridging this gap is profoundly challenging. The dominant paradigms for cultivating advanced reasoning falter here; adapting RL by training a reward model to approximate subjective quality that aligns with human preferences is an immense challenge in itself~\cite{rlhf}, and the subsequent RL process is notoriously sample-inefficient and computationally burdensome~\cite{writingzero}. The alternative, instruction distillation from a powerful model, is often prohibitively expensive and fundamentally capped by the teacher model's capabilities~\cite{openmathinstruct}. This is exacerbated by the scarcity of high-quality queries and deep reasoning trajectories tailored for complex open-ended generation~\cite{longbench}. These constraints create a critical bottleneck, demanding a new paradigm that sidesteps both the sample inefficiency of RL and the costly dependency of distillation.

To break this impasse, we introduce a new paradigm: \textbf{REverse-Engineered Reasoning (REER)}. In contrast to conventional methods that build a reasoning process ``forwards" through trial-and-error or distillation, \emph{we work ``backwards" from a known good outcome}. We essentially ask: ``Given a high-quality piece of output, what is the most coherent and logical thinking process that would have generated it?" By answering this question, we can synthesize the otherwise latent, human-like reasoning paths at scale, bypassing costly distillation of thinking data beforehand or inefficient trial-and-error.

\begin{figure}[t]
    \includegraphics[width=\linewidth]{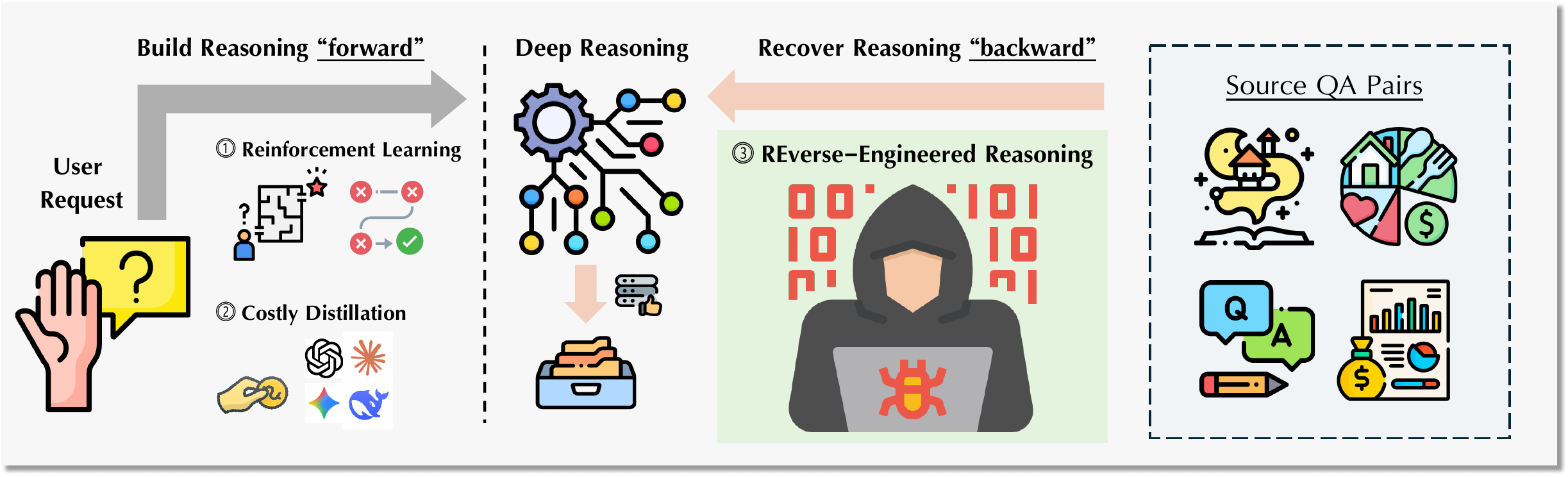}
    \caption{\small \textbf{(Left)} Existing methods attempt to build deep reasoning "forwards" for a user request through trial-and-error (RL) or costly distillation, which falter in open-ended domains that lack clear, verifiable reward signals. \textbf{(Right)} We propose a third path for teaching deep reasoning, REverse-Engineered Reasoning (REER). REER works ``backwards'', recovering plausible human-like thought process from known-good outputs in open-source Question-Answer (QA) pairs.}
    \label{fig:overview}
\end{figure}

We pioneer a novel approach that operationalizes the REER paradigm and, for the first time, \emph{instill deep reasoning capabilities for open-ended generation entirely from scratch}. 
Our approach involves three key stages. First, we source a diverse dataset of query-solution pairs for open-ended generation from the web, encompassing 16,000 samples spanning across ordinary-life question-answering, academic writing, functional writing and creative writing. From these, we ``reverse-engineer" deep reasoning trajectories -- structured, human-like thought process tailored for open-ended generation. 
Eventually, we use this synthetic data to fine-tune a base language model, teaching it to reason and plan deeply before generating a final solution.

The central innovation lies in how we synthesize this data: \emph{we formulate the recovery of high-quality thinking trajectories as a gradient-free search problem.} These trajectories are found by iteratively refining an initial plan, with the search guided by a proxy for thought quality -- the perplexity of a known good solution. The gradient-free, self-contained nature of our synthesis process lends us the scalability. By obviating the need for expensive, query-by-query distillation from proprietary models or the sample-inefficiency of reinforcement learning, our approach provides a cost-effective and automatable pathway to generate vast quantities of high-quality, deep-thinking training data. This makes it possible to instill sophisticated reasoning in models at a scale that was previously impractical.

Using this method, we created \textbf{DeepWriting-20K}, a comprehensive dataset of 20,000 thinking trajectories, and fine-tuned a Qwen3-8B base model. Our extensive empirical evaluation on benchmarks like LongBench~\citep{longbench}, HelloBench~\citep{hellobench}, and WritingBench~\citep{writingbench} shows that DeepWriter-8B successfully internalizes this deep reasoning process. It not only substantially outperforms strong open-source baselines but also achieves performance competitive with, and in some cases exceeding, leading proprietary models like GPT-4o and Claude 3.5, validating our approach as a powerful new pathway for building complex reasoning for open-ended generation.

Our primary contributions are:
\begin{itemize}
    \item \textbf{Pioneering a New Paradigm for Reasoning}: We introduce Reverse-Engineered Reasoning (REER), a novel and scalable paradigm that synthesizes high-quality thinking trajectories without reliance on RL or distillation. This work charts a new, third path for teaching complex reasoning, addressing a critical bottleneck in the field.
    
    \item \textbf{Enabling New Research with DeepWriting-20K:} We contribute a comprehensive dataset of 20,000 query-response and deep reasoning trajectories across 25 categories. This resource is designed to mitigate data scarcity and catalyze future research into the mechanisms of planning and structured thought in open-ended generation.
    \item \textbf{Strong Performance in Open-Ended Generation from Scratch:} We provide rigorous empirical evidence that our approach works. Our model, DeepWriter-8B, trained entirely on our synthesized data, matches or exceeds the performance of premier proprietary models on challenging writing benchmarks, proving that human-like deep reasoning can be cultivated without costly distillation or RL.
\end{itemize}

\def\example{
\begin{figure}[!b]
    \centering
\begin{tcolorbox}[colback=rliableolive!10!white,colframe=black,boxrule=1pt,boxsep=2pt,top=3pt,bottom=3pt,left=2pt,right=2pt]
\textbf{[Question]}

Write an engaging story based on the following writing prompt:

~~``You can't speedrun an Isekai! It's not a game!'' The God tells me.

~~``Watch me.'' I reply.

\textbf{[Response]}

<think>

Alright, the user wants a story based on the prompt ... 

\textcolor{purple}{First, I need to understand the user's intent.} The core objective is to create a compelling ...

Let me think... Maybe I should emphasize the protagonist's journey from disbelief to ...

Hmm... Alternatively, maybe I should also consider the emotional journey of the protagonist and ...

\textcolor{purple}{For the content}, I need to brainstorm a core creative idea. 

The main character, Hiro ... 

\textcolor{purple}{Building the structure}, I'll use a narrative framework with a clear beginning, development, climax. 

Let me think about the beginning  ... Next, the development phase ...
For the climax, I'll need a final, epic battle ... Additionally, I should include Hiro's reflections on his journey ...

Wait, that's a bit too straightforward. 

\textcolor{purple}{Let me add some twists and turns.} 

For instance, the God could ... Hmm, maybe I should include a subplot involving ... This could add layers to the story. ...

</think>

<answer>

...

</answer>

\end{tcolorbox}
\vspace{-.3cm}
    \caption{An example of the deep reasoning generated by DeepWriter. The model demonstrates \textcolor{purple}{deliberate planning} for open-ended generation, employing various thinking patterns (e.g., `Hmm... Alternatively', `Wait, that's a bit ...') to facilitate structured reasoning, including logical deduction, branching, and backtracking.}
    \label{fig:example}
\end{figure}
}
\section{Approach}
\example
Our central goal is to instill deep reasoning in LLMs for open-ended tasks without relying on costly distillation or reinforcement learning. To achieve this, we introduce \textbf{REverse-Engineered Reasoning (REER)}, a novel paradigm that shifts the objective from generating a solution to discovering the latent reasoning process behind an existing high-quality one. Instead of building a reasoning process "forwards" via trial-and-error, REER works "backwards" from a known good output to computationally synthesize the step-by-step thinking that could have produced it. This approach is operationalized as a search problem where we iteratively refine an initial thinking process to discover a trajectory that best explains a high-quality, human-written output. An example of the structured reasoning we aim to cultivate is shown in \textbf{Figure \ref{fig:example}}, where the model demonstrates deliberate planning, exploration of alternatives (``Hmm... Alternatively''), and self-correction (``Wait, that's a bit too straightforward'').

\subsection{REverse-Engineered Reasoning as a Search Problem.}

Let $x$ be an input query (e.g., a story prompt) and $y$ be a high-quality reference solution (e.g., a well-written story). Our objective is to find a \emph{deep reasoning trajectory}, denoted by $z$, which represents a structured, step-by-step thinking process that guides the generation of $y$ from $x$.

The primary challenge in open-ended domains is the absence of a verifiable correctness signal. The REER paradigm circumvents this by reframing the problem: instead of judging the final output, we evaluate the quality of a \textit{thinking process} based on how well it explains a known-good output. We operationalize this principle by using the \textbf{perplexity} (a.k.a, the model surprise) of the reference solution $y$ as a proxy for the quality of a given reasoning trajectory $z$. A lower perplexity score for $y$, conditioned on both $x$ and $z$, indicates that the trajectory provides a more coherent and effective plan. In essence, REER posits that a good thinking process $z$ is one that makes a high-quality answer $y$ seem maximally probable and logical to the model.


Formally, we model the deep reasoning trajectory $z$ as a discrete sequence of reasoning steps, $z = [z_1, z_2, \ldots, z_n]$. The problem is then formulated as a search for the optimal trajectory $z^*$ within the vast space of possible trajectories $\mathcal{Z}$, such that $z^*$ minimizes the perplexity of the reference solution $y$:

$$z^* = \arg\min_{z \in \mathcal{Z}} \text{PPL}(y | x, z)$$

Here, $\text{PPL}(y | x, z)$ is the perplexity of the token sequence of $y$ as calculated by a generator LLM, conditioned on $x$ and $z$. This optimization is performed via a \emph{gradient-free local search algorithm}, allowing us to iteratively refine the trajectory without a differentiable objective.

\subsection{Iterative Refinement via Local Search}
Solving for the optimal trajectory $z^*$ directly is intractable due to the vast search space. Therefore, we propose an iterative refinement algorithm that employs a guided local search to discover a high-quality deep reasoning trajectory. The algorithm starts with an initial trajectory and progressively improves it by refining its constituent segments, guided by the perplexity signal.


\begin{wrapfigure}{R}{0.5\textwidth}
    \includegraphics[width=\linewidth]{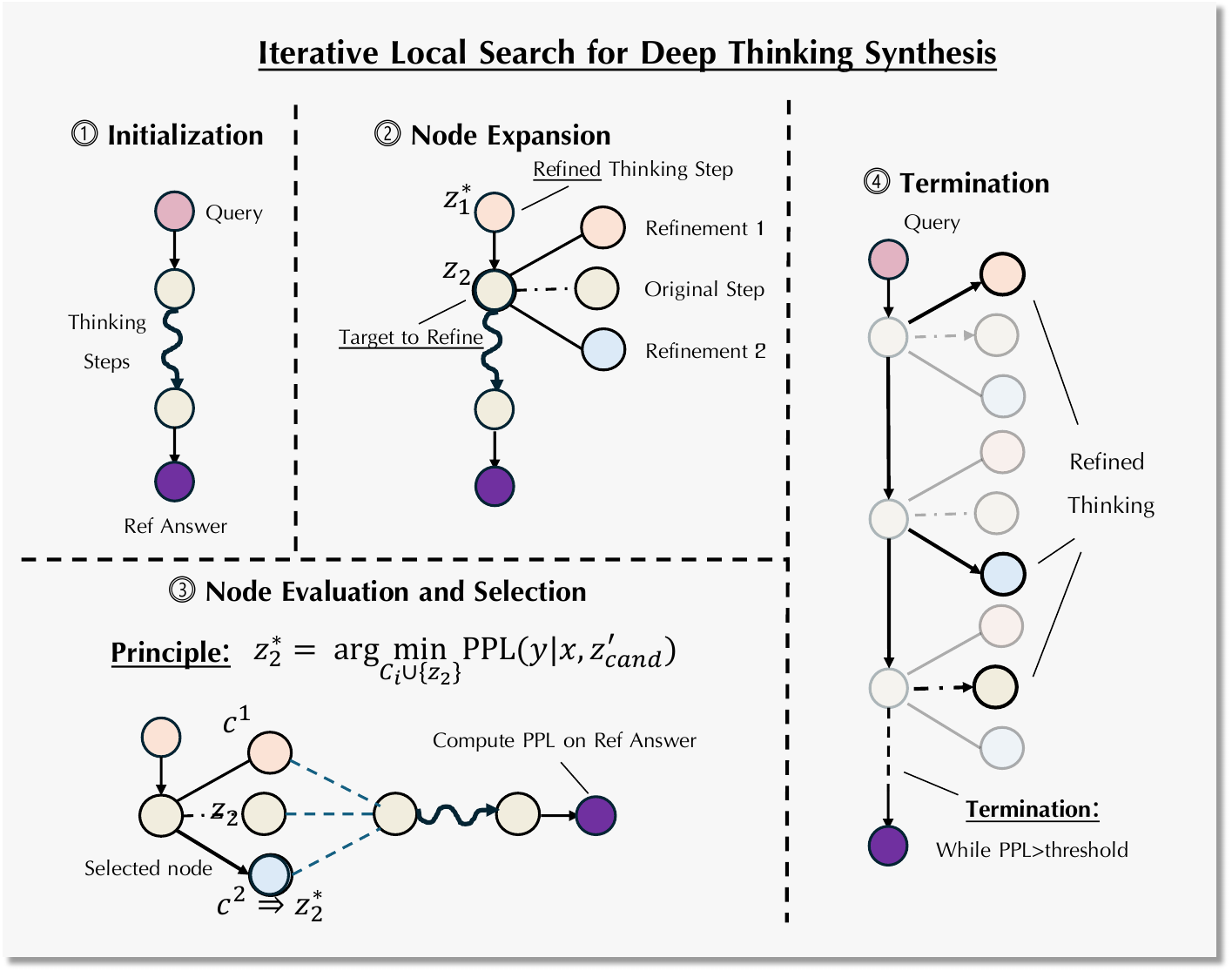}
    \caption{Method Overview: Iterative Local Search for deep reasoning Synthesis. }
    \vspace{-.5cm}
    
    \label{fig:method}
\end{wrapfigure}

The algorithm proceeds as follows, a detailed visualization of this process is shown in \textbf{Figure \ref{fig:method}}.

\textbf{1. Initialization:} For a given $(x, y)$ pair, we generate an initial, imperfect deep reasoning trajectory, $z^{(0)}$, by prompting an LLM with a thought-provoking instruction (see Appendix, Listing 1) to produce a plausible plan. This initial trajectory is denoted as $z = [z_1, z_2, \ldots, z_n]$.

\textbf{2. Node Expansion (Segment-wise Refinement):} The core of our method is an iterative loop that refines $z$ one segment at a time. In each iteration, we select a segment $z_i$ to improve. We prompt the LLM to generate candidate refinements with more thinking-based details, elaborations and reflections. To generate these refinements, we provide the full context including the query $x$, the reference solution $y$, and the surrounding trajectory segments (refined steps $z_{<i}^\ast$ and initial steps $z_{>i}$). The prompt is meticulously designed to encourage detailed reasoning while preventing the model from simply copying content from the reference solution (see Appendix, Listing 2).

\textbf{3. Node Evaluation and Selection:} For each generated candidate $c$, we construct a temporary trajectory $z'_{\text{cand}}$ by substituting $z_i$ with $c$. We then evaluate each candidate by computing its quality score, $S(c) = \text{PPL}(y | x, z'_{\text{cand}})$. The candidate with the lowest perplexity score is chosen as the updated segment for the next iteration: $z_i^{*} = \arg\min_{c \in \mathcal{C}_i \cup \{z_i\}} S(c)$. We include the original segment $z_i$ in the candidate set to ensure that the perplexity improves monotonically.

\textbf{4. Termination:} The refinement process repeats until the perplexity of the solution reaches a predefined target threshold or a maximum number of iterations is completed. The final output is a refined trajectory $z^*$.

This process allows us to create a dataset of $(x, z^*, y)$ triples, which can then be used to fine-tune a base LLM to internalize the deep reasoning capability for open-ended generation from scratch. 

It is important to distinguish our iterative local search from methods like Monte Carlo Tree Search (MCTS)~\cite{mcts,fastmcts}. First, by using the perplexity of a complete reference solution as a quality proxy, REER avoids the computationally expensive rollouts required in MCTS. Second, our approach operates on a "global-to-local" principle: we start with a complete, albeit imperfect, global plan  and iteratively improve it through local, segment-wise edits. This contrasts with standard MCTS or beam search, which build solutions sequentially by extending partial states. These distinctions make our approach a scalable and efficient method for operationalizing REER, enabling the creation of large-scale, deep-reasoning datasets for open-ended generation.


\subsection{Training Data Curation}
The success of our methodology hinges on a large-scale, high-quality dataset of $(x, z^*, y)$ triples. The creation of this dataset follows a multi-stage pipeline: sourcing diverse query-solution pairs, synthesizing deep reasoning trajectories, and applying rigorous filtering.

\subsubsection{Sourcing of (Query, Solution) Pairs}
To ensure diversity in style, topic, and complexity, we sourced initial $(x, y)$ pairs from three primary channels:
\begin{itemize}
    \item \textbf{Public Writing Platforms:} We scraped prompt-story pairs from communities like `r/WritingPrompts`, using community upvotes as an initial quality proxy.
    \item \textbf{Public Domain Literature:} We used classic texts from Project Gutenberg as solutions ($y$) and prompted GPT-4o to reverse-engineer plausible queries ($x$) from their opening paragraphs.
    \item \textbf{Public Datasets:} We augmented our collection with data from instruction tuning datasets such as WildChat~\citep{wildchat} and LongWriter6K~\cite{longbench}.
\end{itemize}

\subsubsection{Trajectory Synthesis and Filtering}
From our sourced pairs, we selected 20,000 high-quality query-solution pairs covering 25 manually nominated categories to ensure broad topic coverage. For each pair, we executed our iterative local search algorithm to generate an optimal deep reasoning trajectory $z^*$. 

\textbf{Context Engineering.}
The efficacy of the search algorithm, however, hinges not only on the search procedure but also on the nuanced design of the instructions used to elicit deep reasoning from the generator LLM. We proposed three key designs in our context engineering to ensure high-quality synthesis. We only summarize the key insights here and refer the reader to the appendix for detailed prompts.

\begin{enumerate}

    \item \textbf{Enforcing Segment-wise Edits via a Meta-Structure.}
To ensure the generator model performs a true segment-wise edit without including edits for the subsequent parts of the trajectory, we enforce a \emph{meta-structure} for the reasoning process within the prompt. This serves as an \emph{implicit regularizer}, helping the model to localize the current segment and constrain its edits to the intended scope when performing segment-wise edits.
\item \textbf{Injecting Human-like Thinking Patterns.}
To prevent the synthesis of rigid and formulaic reasoning, we \emph{deliberately inject human-like thinking patterns}. Prompts explicitly encourage phrases that signify cognitive exploration and self-reflections (e.g., ``Hmm...maybe I can...'', ``Wait, that's a bit...''), triggering a more human-like reasoning style and incentivizing self-reflection through training~\cite{wang2025vl}.
\end{enumerate}

\begin{figure}[h]
    \centering
\includegraphics[width=1.0\textwidth]{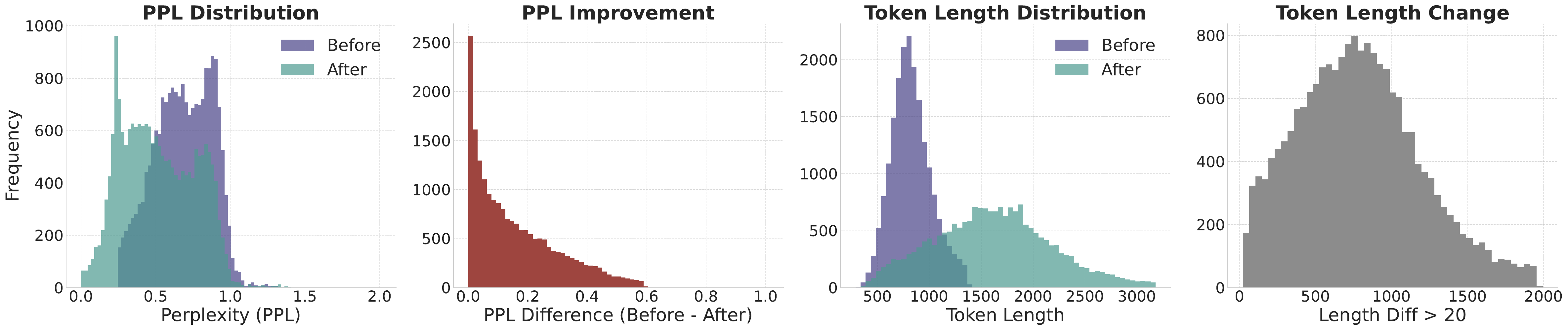}
    \caption{Analysis of Token Length \& Perplexity Before and After the Search. The leftmost two plots show that our iterative search process consistently \textbf{reduces perplexity (PPL)}. The rightmost two plots show that the process also tends to \textbf{increase the token length} of the thinking trajectory, reflecting the addition of more detailed reasoning steps.}
    \label{fig:search_analysis}
\end{figure}

Analysis of this synthesis process, shown in \textbf{Figure \ref{fig:search_analysis}}, confirms its effectiveness. The perplexity distribution shifts significantly lower after refinement, with the vast majority of samples showing a marked PPL improvement. Concurrently, the token length of the trajectories increases, to an  indicating that the search process successfully expands initial simple plans into more detailed and elaborate reasoning chains.

During instruction tuning, we witness the challenge of repetitive and degenerate thinking. We therefore applied two heuristic filtering strategies to prune low-quality trajectories:
\begin{enumerate}
    \item \textbf{End-of-Thinking Filtering:} We discarded samples where thinking patterns persisted in the final 10\% of the sequence. These trajectories risk misleading the model to stuck in a repetitive loop and failing to conclude its reasoning process.
    \item \textbf{Repetition Filtering:} We employed a repetition metric to measure the frequency of the top-3 n-grams within each trajectory. Samples exhibiting high n-gram repetition, a sign of degenerative looping expressions, were filtered out.
\end{enumerate}

\begin{figure}[h!]
\centering
\includegraphics[width=0.8\textwidth]{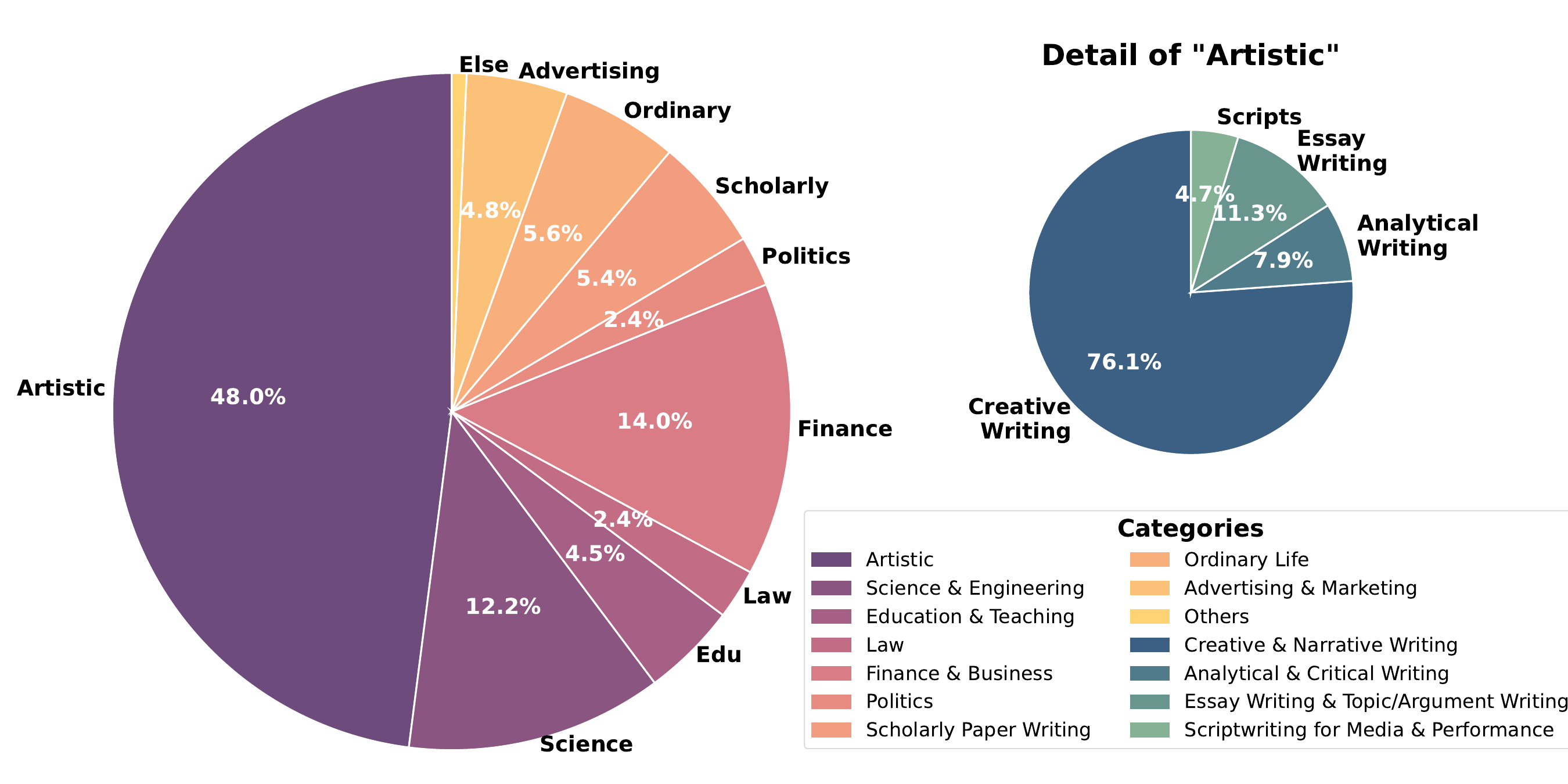}
\caption{Distribution of the final 20K training dataset by categories taking more than 0.5\% account. The primary chart shows a diverse range of topics, with a large emphasis on \textbf{Artistic} writing. The detailed view of "Artistic" reveals a focus on Creative Writing and other styles, ensuring comprehensive coverage for open-ended generation.}
\label{fig:data_pie}
\end{figure}

This process resulted in a final dataset of 20,000 high-quality deep reasoning trajectories. The distribution of this dataset, shown in \textbf{Figure \ref{fig:data_pie}}, highlights its diversity, with a significant focus on \textbf{Artistic} (Literature and Arts) writing, which is further broken down into sub-genres like Creative Writing and Essay Writing.

\subsubsection{Final Dataset Assembly for Fine-Tuning}
Training a model exclusively on domain-specific data risks overfitting and can degrade its general knowledge priors. To mitigate this, we adopted a \textbf{mixed-data training strategy}. We combined our 20K synthetically generated trajectories with distilled deep reasoning trajectories from public datasets (OpenThoughts~\cite{guha2025openthoughts}) that primarily cover domains like mathematics, coding, and science. This blended datasets prevents the model from catastrophic overfitting when learning deep reasoning for open-ended generation.

To train the base LLM, each complete triple in the final dataset was formatted using the prompt template shown in Listing 4 in the appendix. This structure explicitly teaches the model to first perform deep reasoning before producing the final output, thereby internalizing the desired reasoning process from scratch.

\section{Experiments}

Our empirical evaluation is structured to rigorously validate the efficacy of DeepWriter. We address two central research questions:
\begin{enumerate}
    \item How does DeepWriter, fine-tuned from scratch on an 8B open-source model, compare against state-of-the-art proprietary models and other powerful open-source alternatives across a spectrum of diverse open-ended generation tasks?
    \item What is the individual contribution of each core component of our approach -- specifically, the synthesized deep thinking trajectories, the iterative refinement algorithm, and the characteristics of the thinking traces and the data composition  -- to the model's final performance?
\end{enumerate}
To answer these questions, we first present a comprehensive comparison against leading models, followed by a series of targeted ablation studies and a qualitative deep-dive analysis into the model's reasoning capabilities.

\subsection{Experimental Setup}

\begin{itemize}
    \item \textbf{Training Data:} Our primary training dataset comprises approximately 20,000 deep thinking trajectories, which we synthesized for 16,000 unique queries spanning a wide array of open-ended tasks. As stated in Section 3, to prevent catastrophic forgetting of general reasoning abilities, we blended this core dataset with public thinking-process datasets that reasoning-related domains (e.g., mathematics, coding). This resulted in a final mixed dataset of 37,000 examples, ensuring a balance between specialized open-ended generation capabilities and keeping broad knowledge priors.
    
    \item \textbf{Implementation Details:} We selected \textbf{Qwen3-8B-Base} as our base model for fine-tuning. This decision was informed by preliminary experiments where other candidates, such as Llama-3.1-8B-Base, struggled to effectively internalize the deep thinking process, and Qwen-2.5-7B-Base faced prohibitive context length limitations. For the critical trajectory synthesis stage, we utilized Qwen2.5-32B-Instruct as the generator model. Fine-tuning was conducted for 3 epochs using a constant learning rate schedule, with a peak learning rate of $2 \times 10^{-5}$ and a global batch size of 96.
\end{itemize}

\subsection{Evaluation Benchmarks}
To ensure a comprehensive and multi-faceted evaluation, we employed a suite of three complementary benchmarks: LongBench-Write (LB), HelloBench (HB), and WritingBench (WB). Together, they probe three distinct and critical dimensions of generative performance: raw endurance, real-world applicability, and domain-specific proficiency.

\begin{itemize}
    \item \textbf{LongBench-Write (LB):} This benchmark functions as a targeted stress test for generative endurance. It is designed to measure a model's ability to produce coherent, ultra-long-form text (e.g., >10,000 words), allowing us to assess the foundational capacity for maintaining thematic consistency over extended outputs.
    
    \item \textbf{HelloBench (HB):} To gauge practical applicability, HelloBench evaluates performance on a diverse set of "in-the-wild" tasks sourced from real user queries. Our analysis focuses on two key subsets: \textbf{HB-A (Open-Ended QA)}, which tests the generation of detailed and nuanced answers, and \textbf{HB-B (Heuristic Text Generation)}, which assesses creative reasoning and stylistic fidelity in long-form narrative continuation.
    
    \item \textbf{WritingBench (WB):} This benchmark is tailored to measure domain-specific proficiency and controllability across six professional and creative domains: \textbf{A} (Academic \& Engineering), \textbf{B} (Finance \& Business), \textbf{C} (Politics \& Law), \textbf{D} (Literature \& Arts), \textbf{E} (Education), and \textbf{F} (Advertising \& Marketing). It specifically evaluates the ability to adhere to complex, multi-dimensional constraints, a hallmark of advanced open-ended generation.
\end{itemize}

\noindent\textbf{Evaluation Protocols:} Given the subjective nature of open-ended tasks, we adopted the established protocol  of using powerful LLMs as judges within each benchmarks\footnote{We adopted the latest evaluation protocol of WritingBench, which, however, resulted a discrepancy in performance compared with the original WritingBench paper. This discrepancy is also acknowledged by the authors.}. While we acknowledge the potential for inherent biases in this method, it remains the most scalable and consistent approach for evaluating nuanced generative quality. Specifically, \textbf{Claude-3.7} was used to score outputs for LongBench and WritingBench, while \textbf{GPT-4o} was used for HelloBench.

\subsection{Main Results}
We benchmarked DeepWriter against leading proprietary models (GPT-4o, Claude 3.5, Claude 3.7) and a strong open-source baseline, LongWriter-8B. The results, presented in Table~\ref{tab:main_results}, unequivocally demonstrate that our methodology successfully instills sophisticated generation capabilities in an 8B model from scratch.

\begin{table}[h!]
\centering
\caption{Main performance comparison on LongBench (LB), HelloBench (HB), and WritingBench (WB). DeepWriter, an 8B model fine-tuned from scratch, demonstrates competitive performance against leading proprietary models and significantly outperforms other open-source models in its class.}
\label{tab:main_results}
\begin{tabular}{lcccccccccc}
\toprule
\textbf{Model} & \textbf{Base Model} & \textbf{LB} & \textbf{HB-A} & \textbf{HB-B} & \textbf{WB-A} & \textbf{WB-B} & \textbf{WB-C} & \textbf{WB-D} & \textbf{WB-E} & \textbf{WB-F} \\
\midrule
GPT-4o & - & 83.1 & 83.7 & 87.6 & 74.40 & 73.42 & 74.38 & 77.91 & 75.86 & 78.08 \\
Claude 3.5 & - & 89.3 & 82.9 & 88.3 & 59.05 & 57.68 & 56.32 & 59.36 & 62.00 & 67.70 \\
Claude 3.7 & - & 97.8 & 83.9 & 93.2 & 78.24 & 77.93 & 76.51 & 79.37 & 79.26 & 80.88 \\
\midrule
LongWriter-8B & Llama3.1-8b & 76.5 & 80.1 & 82.6 & 57.97 & 53.92 & 49.08 & 52.08 & 52.99 & 52.08 \\
\textbf{DeepWriter-8B} & \textbf{Qwen3-8b} & \textbf{91.28} & \textbf{82.64} & \textbf{87.48} & \textbf{72.20} & \textbf{71.76} & \textbf{70.57} & \textbf{70.57} & \textbf{73.65} & \textbf{72.29} \\
\bottomrule
\end{tabular}
\end{table}

\textbf{Analysis of Main Results.} The results in Table~\ref{tab:main_results} reveal several compelling findings. First, \textbf{DeepWriter-8B consistently and substantially outperforms the strong open-source baseline, LongWriter-8B, across all benchmarks}. The performance gap is particularly stark in the diverse WritingBench domains, where DeepWriter achieves an average uplift of over 18 points. This highlights the profound advantage of our deep thinking synthesis approach over standard instruction tuning for cultivating advanced generative skills.

Second, \textbf{DeepWriter-8B closes a significant portion of the performance gap with elite proprietary models}. On the creative HelloBench task (HB-B), its score (87.48) is statistically on par with GPT-4o (87.6) and Claude 3.5 (88.3). More strikingly, on the professional writing tasks in WritingBench, DeepWriter-8B not only surpasses Claude 3.5 by a large margin in all six categories but also remains highly competitive with the much larger GPT-4o and Claude 3.7 models. A counter-intuitive result is DeepWriter-8B's score of \textbf{91.28 on LongBench-Write}, exceeding both GPT-4o (83.1) and Claude 3.5 (89.3). This suggests that explicitly training on structured thinking trajectories provides a powerful inductive bias for maintaining long-range coherence, a critical challenge in ultra-long text generation.

\subsection{Ablation Studies}
To meticulously dissect the contribution of each component of our methodology, we conducted a series of ablation studies, with results detailed in Table~\ref{tab:ablation_results}. Each experiment isolates a specific design choice to quantify its impact on overall performance.

\begin{table}[h!]
\centering
\caption{Ablation studies. The full model (top row) is compared against versions with key components removed. Results show that the synthesized deep thinking trajectories and iterative refinement are crucial for performance.}
\label{tab:ablation_results}
\resizebox{\textwidth}{!}{%
\begin{tabular}{lccccccccc}
\toprule
\textbf{Model Configuration} & \textbf{LB} & \textbf{HB-A} & \textbf{HB-B} & \textbf{WB-A} & \textbf{WB-B} & \textbf{WB-C} & \textbf{WB-D} & \textbf{WB-E} & \textbf{WB-F} \\
\midrule
\textbf{DeepWriter-8B (Full)} & \textbf{91.28} & \textbf{82.64} & \textbf{87.48} & \textbf{72.20} & \textbf{71.76} & \textbf{70.57} & \textbf{70.57} & \textbf{73.65} & \textbf{72.29} \\
\midrule
- Remove Synthesis Data & 82.93 & 70.92 & 73.73 & 63.44 & 62.78 & 62.86 & 57.72 & 66.32 & 62.78 \\
- Remove Iterative Search & 83.20 & 81.08 & 84.48 & 66.72 & 68.79 & 67.36 & 65.66 & 69.53 & 70.13 \\
- Remove Reflection Tokens & 86.97 & 82.27 & 82.80 & 71.68 & 69.64 & 70.44 & 62.04 & 69.98 & 71.94 \\
- Downsample Long Traces & 90.38 & 82.26 & 84.07 & 69.63 & 70.36 & 69.10 & 67.52 & 69.84 & 70.70 \\
- Downsample Short Traces & 89.34 & 81.12 & 82.19 & 70.89 & 70.60 & 70.04 & 66.93 & 72.40 & 69.75 \\
- Remove Literature \& Arts data & 88.82 & 81.66 & 85.32 & 71.37 & 71.02 & 69.33 & 69.80 & 72.20 & 71.30 \\
\bottomrule
\end{tabular}
}
\end{table}

The ablation results provide robust evidence supporting our methodological design.

\begin{itemize}
    \item \textbf{Importance of Synthesized Data:} Removing our 20K synthesized trajectories and training only on public thinking datasets ("- Remove Synthesis Data") causes the most significant performance degradation across the board. Scores plummet, particularly in creative tasks like HelloBench HB-B ($87.48 \rightarrow 73.73$) and across WritingBench (average drop of over 8 points). This confirms a core hypothesis: it is not merely the presence of "thinking" data that matters, but the \textbf{quality and relevance of structured trajectories tailored for open-ended domains} that drive performance.

    \item \textbf{Impact of Iterative Refinement:} Using the initial, unrefined thinking trajectories ($z^{(0)}$) instead of the final, optimized ones ($z^{*}$) ("- Remove Iterative Search") also leads to a clear drop in performance. While the decline is less severe than removing the synthesis data entirely, the drop on nuanced WritingBench tasks (e.g., WB-A: $72.20 \rightarrow 66.72$) is substantial. This proves that our perplexity-guided local search is highly effective at discovering superior reasoning paths that translate directly into stronger generative capabilities.

    \item \textbf{Effect of Reflection Tokens:} Removing reflection tokens (e.g., 'Hmm...', 'Wait, that's...') from the synthesis prompts ("- Remove Reflection Tokens") has a nuanced effect. While overall scores dip slightly, the most pronounced drop is in WritingBench domain D (Literature \& Arts), which falls from 70.57 to 62.04. This suggests that these explicit markers of cognitive exploration, self-correction, and branching are particularly valuable for instilling the flexibility and creativity required in artistic writing tasks.

    \item \textbf{Role of Trajectory Length:} We explored the impact of trace length by selectively downsampling either long or short trajectories. The results reveal a task-dependent preference: removing longer, more elaborate traces ("- Downsample Long Traces") disproportionately harms performance on complex, domain-specific tasks like those in WritingBench. Conversely, removing shorter, more concise traces ("- Downsample Short Traces") has a slightly larger negative impact on creative tasks like HB-B. This suggests that detailed, multi-step plans are crucial for structured professional writing, while nimbler, more direct reasoning may be optimal for creative ideation.

    \item \textbf{Role of Literature \& Arts Data:} Removing the data from the "Literature \& Arts" and "Ordinary Life" domains ("- without Literature \& Arts data") degrades performance across all benchmarks, not just in the corresponding WB-D category. This finding indicates that training on creative and narrative tasks imparts a more generalizable ability to handle nuance, structure, and open-endedness, even benefiting performance in more technical domains. This highlights the contribution of the release of our 20K dataset covering comprehensive topics. 
\end{itemize}

\subsection{Qualitative Analysis}
\subsubsection{Generation 
Quality of Deep Thinking}
\begin{wrapfigure}{r}{0.4\textwidth}
    \includegraphics[width=1.0\linewidth]{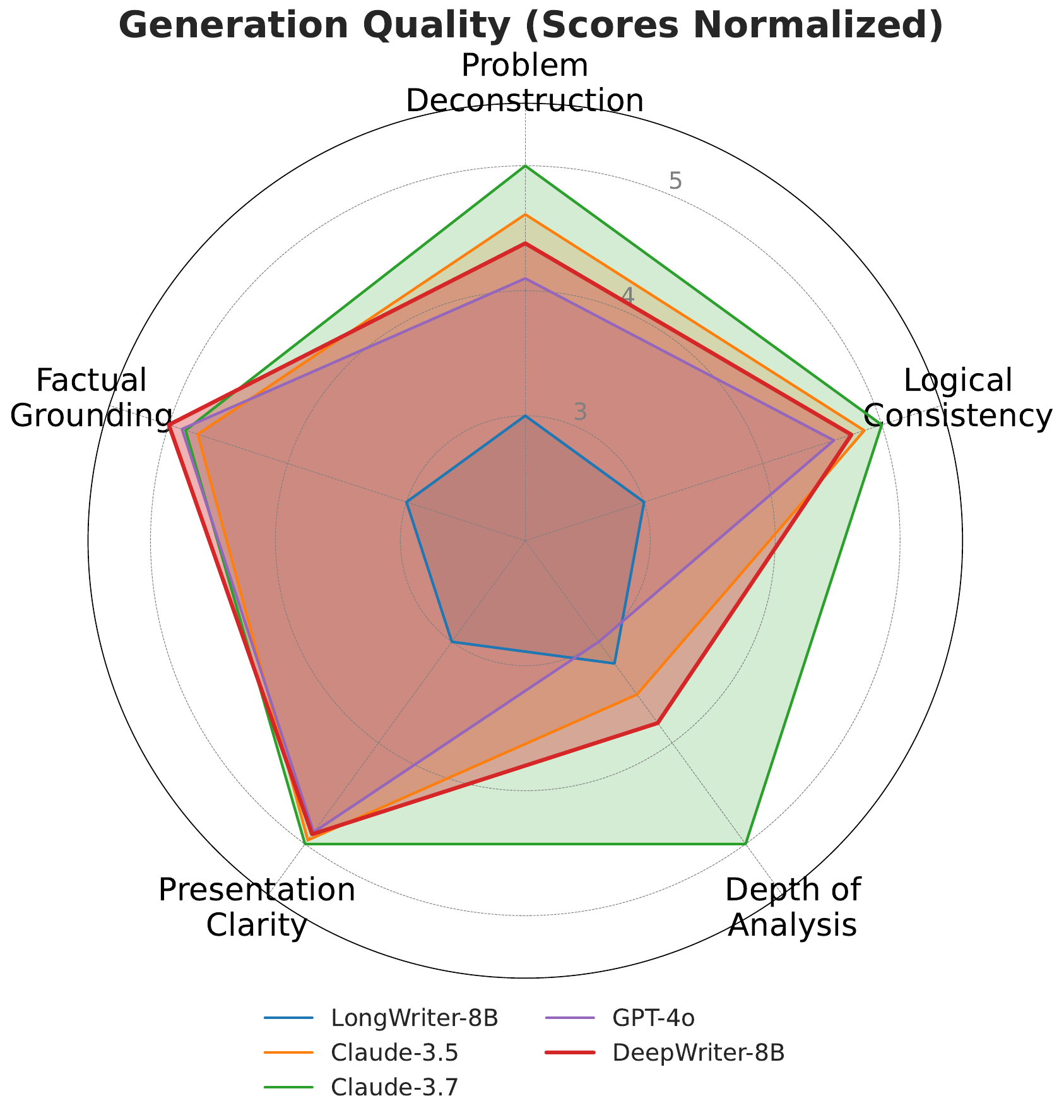}
    \caption{Qualitative comparison of generation quality. Scores are normalized across five dimensions related to deep thinking. DeepWriter-8B shows a reasoning profile far superior to the open-source baseline and is competitive with top proprietary models.}
    \label{fig:radar_chart}
    \vspace{-1cm}
\end{wrapfigure}
Beyond quantitative scores, we sought to understand how well DeepWriter internalizes the \textit{qualities} of deep thinking. To this end, we conducted a qualitative analysis, scoring model outputs on five dimensions intrinsically linked to advanced reasoning and planning:

\begin{itemize}
    \item \textbf{Problem Deconstruction:} The ability to break down a complex prompt into a logical hierarchy of sub-goals. This is the foundation of effective planning.
    \item \textbf{Logical Consistency:} Maintaining a coherent and non-contradictory reasoning path throughout the entire generation necessistates the ability to plan over the generation.
    \item \textbf{Depth of Analysis:} Moving beyond surface-level responses to explore nuances, consider alternatives, and demonstrate sophisticated understanding. This reflects the "deep" aspect of the thinking process.
    \item \textbf{Presentation Clarity:} The ability to structure the final output in a clear, organized, and persuasive manner, which is a direct outcome of a well-formed internal plan.
    \item \textbf{Factual Grounding:} Ensuring that generated content, where applicable, is accurate and well-supported, reflecting a robust and reality-aware reasoning process.
\end{itemize}

The normalized scores, visualized in the radar chart in Figure~\ref{fig:radar_chart}, provide a signature of each model's reasoning profile. As illustrated in Figure~\ref{fig:radar_chart}, DeepWriter-8B exhibits a remarkably strong and well-rounded reasoning profile. Its performance polygon significantly envelops that of the LongWriter-8B baseline, showing dramatic improvements across all five dimensions. This confirms that our methodology genuinely enhances underlying reasoning capabilities, rather than just improving superficial output fluency.

Furthermore, DeepWriter-8B's profile closely rivals that of GPT-4o and substantially exceeds Claude 3.5, particularly in \textbf{Depth of Analysis} and \textbf{Factual Grounding}. While the state-of-the-art Claude 3.7 still defines the frontier, especially in Depth of Analysis, our 8B model has demonstrably bridged a large portion of the capability gap. This validates our central claim: instilling a deep thinking process through gradient-free synthesis is a highly  promising pathway toward building more powerful and scalable models.



\subsubsection{Qualitative Comparison of Thinking Patterns}

To better understand how injecting human-like thinking patterns during synthesis affects the model's behavior, we analyze the frequency of reasoning phrases generated by the full model versus the ablated model trained without injecting thinking patterns. The thinking patterns are deduplicated such that occurances will be counted only once for the same solution.

\begin{figure}[h!]
    \centering
    \includegraphics[width=1.0\linewidth]{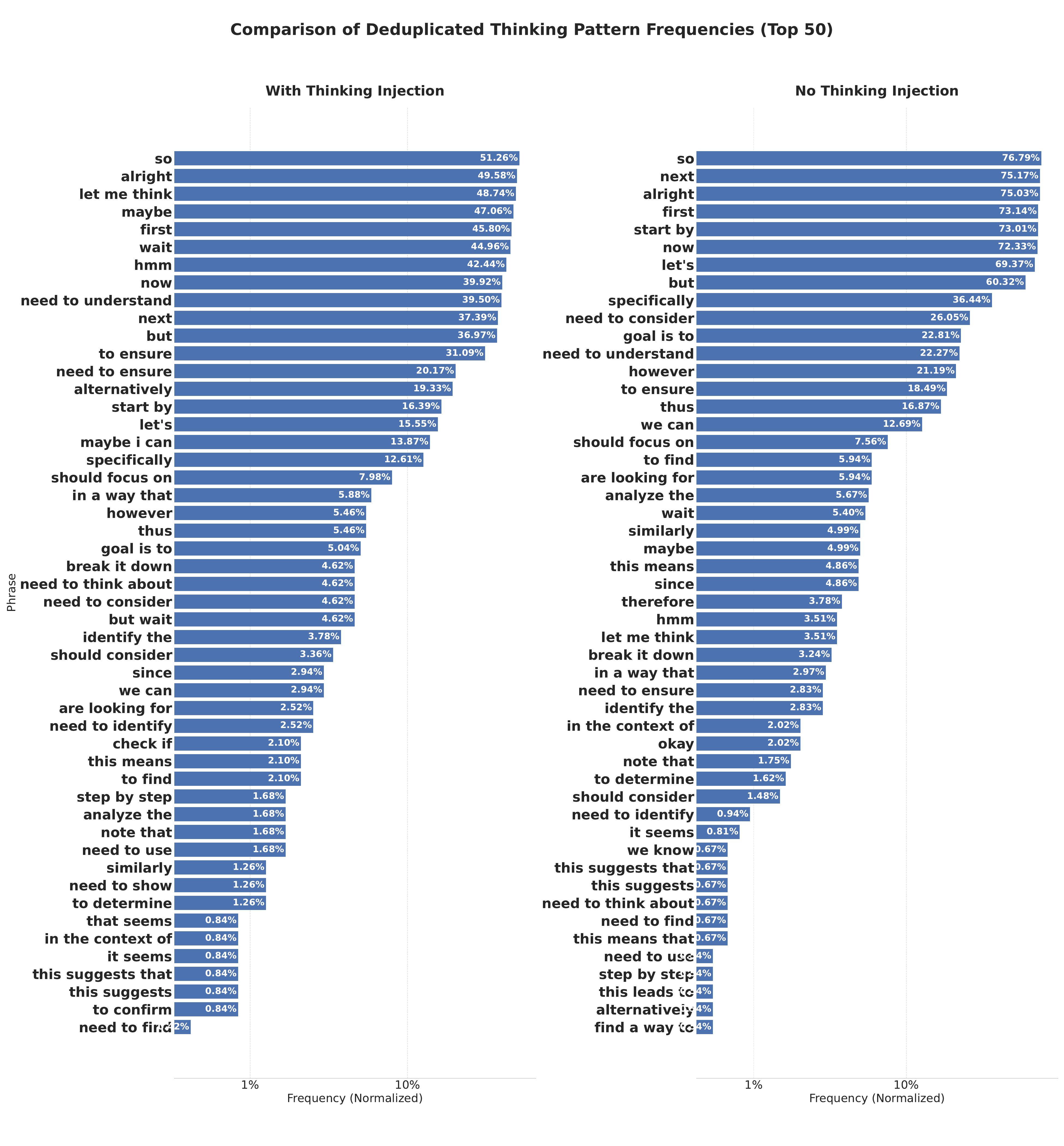}
    \caption{Comparison of the top 50 thinking pattern frequencies for models trained with and without the injection of human-like thinking patterns during data synthesis. The model with injection (left) shows a more diverse and balanced distribution of patterns, while the model without (right) relies heavily on a few formulaic phrases.}
    \label{fig:thinking_patterns}
\end{figure}

As shown in Figure~\ref{fig:thinking_patterns}, the difference is stark. The model trained \textit{with} thinking pattern injection exhibits a more diverse and evenly distributed use of thinking patterns. Tokens indicating reflection and self-correction, such as `let me think', `maybe', `hmm', and `wait', are prominent. This suggests a more flexible, human-like reasoning process with cognitive exploration. In contrast, the model trained \textit{without} this injection relies on a small set of highly frequent phrases like `next', `first', and `goal is to'. The frequency distribution is highly skewed, indicating a more rigid and formulaic reasoning process.

This analysis confirms that the proposed context engineering techniques encourages the model to adopt a more nuanced and reflective approach to problem-solving, which, as shown in the ablation studies, is particularly beneficial for creative and complex tasks.
\section{Related Work}

\textbf{Deep Reasoning and Test-Time Computation.} The paradigm of ``deep reasoning'' (or Long CoTs) aims to move beyond rapid, surface-level inference by leveraging increased computational investment at test time. Advanced models from organizations like Google \citep{gemini}, DeepSeek AI \citep{deepseek}, and OpenAI \citep{oai} have demonstrated the effectiveness of this test-time scaling~\cite{qwq, s1, deepthink}. This approach gained prominence with methods like Chain-of-Thought (CoT) prompting \citep{cot}, which elicits intermediate reasoning steps to guide a model toward more accurate solutions. Building on this, more sophisticated strategies have emerged, such as Tree-of-Thought (ToT) \citep{tot}, which explores a tree of possible reasoning paths, and various self-correction or self-refinement~\citep{madaan2023self,kumar2024training, star, qstar} mechanisms that iteratively improve an initial response. While these approaches have yielded remarkable performance gains in verifiable domains like mathematics and programming, their application to open-ended, creative tasks remains largely unexplored due to the absence of a singular ground truth for verification. REER addresses this gap by developing a method to instill this deliberate, structured thinking capability for non-verifiable creative domains.

\textbf{Paradigms for Instilling Reasoning.} Beyond prompting techniques at inference time, two dominant paradigms exist for integrating advanced reasoning capabilities directly into a model's parameters: reinforcement learning and instruction distillation.

Reinforcement Learning (RL) has been instrumental in aligning LLMs with human preferences (RLHF) and improving performance on tasks with clear reward signals~\cite{rlhf, deepseek, kimi-1_5, wang2025emergent}. In verifiable domains, a correct outcome provides a straightforward positive reward, effectively guiding the model's search through a vast solution space \citep{dsmath, wang2025code, wang2025vl, su2025pixel}. However, this reliance on verifiability presents a formidable barrier when applied to open-ended generation~\cite{rlhf, writingzero}. Crafting a reward model that can reliably approximate nuanced and subjective qualities like originality or emotional resonance is an immense challenge in itself \citep{rlhf, genrm}. Furthermore, the subsequent RL process is often computationally burdensome and sample-inefficient~\cite{dsmath,gulcehre2023rest, wang2023adversarial}. Recently VeriFree~\cite{verifree} extends verification-based reward to likelihood-based reward for reinforcement learning on verifiable domains. Likewise, REverse-Engineered Reasoning (REER) shares the principle of using a proxy to judge the reasoning quality. However, the motivation is fundamentally different -- we focus on recovering human-like deep reasoning from known-good outputs for the broader open-ended generation problems.

Instruction Distillation offers an alternative, wherein reasoning traces are generated by a powerful ``teacher'' model (e.g., GPT-4 \citep{gpt4}) and used as training data for a smaller ``student'' model. While effective, this approach is constrained by two fundamental limitations. First, it is often hampered by the prohibitive cost of querying state-of-the-art proprietary models at scale \citep{guha2025openthoughts, openmathinstruct}. Second, and more fundamentally, distillation is capped by the teacher's abilities—a student model cannot learn a capacity that the teacher does not already possess~\cite{openmathinstruct}. This limitation is exacerbated by the general scarcity of high-quality, open-source instruction data tailored for advanced creative tasks~\cite{longbench}.

To overcome these data bottlenecks, researchers have increasingly turned to synthetic data generation. Most approaches use a powerful LLM to generate new query-response pairs, often to augment existing datasets or bootstrap capabilities in new domains~\citep{selfinstruct, star, qstar, gu2025realsyn, Yang_2023_ICCV, han2025attributestextualgenesleveraging}. These methods aim to build a solution ``forwards'' for a given query through data synthesis. Our central innovation is to "reverse-engineer" reasoning -- synthesize deep reasoning ``backwards'' from a known good outcome such as human-written solutions. 

\textbf{Writing Datasets, Models and Benchmarks}
Prior work has explored both synthetic data pipelines and RL in AI writing. For instance, Weaver~\citep{wang2024weaver} proposed instruction back-translation, LongWriter \citep{longbench} proposed an agentic data pipeline to synthesize long-form writing outputs and introduced the LongBench-Write benchmark. In contrast, Writing-Zero \citep{writingzero} employed an RL approach, training a reward model on private datasets, but its training data remains unreleased. DeepWriter, to our knowledge, is the first to \emph{instill deep reasoning for open-ended generation} using a scalable, open synthetic data approach.

Evaluation in this domain relies on recently developed benchmarks. HelloBench \citep{hellobench} proposes a diverse collection of "in-the-wild" tasks from real user queries to gauge practical applicability. Meanwhile, WritingBench \citep{writingbench} measures domain-specific proficiency and the ability to adhere to complex, multi-dimensional constraints across six professional domains.

\section{Conclusion}
This work addresses the critical challenge of instilling deep reasoning capabilities in Large Language Models for open-ended, non-verifiable domains -- a frontier where dominant paradigms like reinforcement learning and instruction distillation have faltered due to the absence of clear reward signals and the prohibitive costs of distillation. We introduced Reverse-Engineered Reasoning (REER), a novel paradigm that fundamentally shifts the approach, from building reasoning ``forwards'' through trial-and-error or distillation, to recovering the latent, step-by-step thinking process ``backwards'' using known high-quality outputs. By framing this recovery as a gradient-free local search problem guided by the perplexity, \emph{REER provides a scalable, cost-effective, and automatable "third path" for cultivating sophisticated reasoning without relying on RL or expensive teacher models.} This new paradigm results in the creation of the DeepWriting-20K dataset, a large-scale, open-source collection of 20,000 deep reasoning trajectories for open-ended tasks. \emph{We release the high-quality dataset to address the data scarcity  and facilitate further research in open-ended generation}. Trained from scratch on this data, DeepWriter-8B achieves strong performance across benchmarks, \emph{providing compelling empirical evidence that human-like deep reasoning can be successfully cultivated in smaller models, democratizing access to capabilities previously confined to large-scale, proprietary systems.}

\clearpage

\bibliographystyle{plainnat}
\bibliography{main}

\clearpage

\beginappendix

\section{List of Prompts}
Below we list the exact prompts used for trajectory synthesis and in-house evaluation. For the meta-structure guidelines and thinking pattern injection, refer to Listing~\ref{lst:initial}. For enforcing segment-wise edits, refer to Listing~\ref{lst:edits}. For quality assessment with regard to deep reasoning, refer to Listing~\ref{lst:quality}. For computing the proxy score of a deep reasoning trajectory, we employ Listing~\ref{lst:infer}, without including the reference output. 
\lstlistoflistings

\begin{lstlisting}[style=rawwrap,
  caption={Prompt for Generating Initial Thinking.}, % This is the caption text
  label={lst:initial}]
You are an expert in many fields. Suppose you will give a specific final response, I need you to also write down the thought process behind this solution.
Here is a task:
{}

Here is the solution you will create:
{}

Now, you need to write down the thinking process behind this solution, as if you are thinking aloud and brainstorming in the mind. The thinking process involves thoroughly exploring questions through a systematic long thinking process. This requires engaging in a comprehensive cycle of analysis, summarizing, exploration, reassessment, reflection, backtracing, and iteration to develop well-considered thinking process. Present your complete thought process within a single and unique `<think></think>` tag.

Your thought process must adhere to the following requirements:

1.  **Narrate in the first-person as if you are thinking aloud and brainstorming**
    Stick to the narrative of "I". Imagine you are brainstorming and thinking in the mind. Use verbalized, simple language.

2.  **Unify the thinking process and the writing solution:**
    Your thought process must precisely correspond to a part of the writing solution. The reader should be able to clearly see how your thoughts progressively "grew" into the finished piece, making the copy feel like the inevitable product of your thinking.

3.  **Tone of Voice: Planning, Sincere, Natural, and Accessible**
    Imagine you are analyzing and planning what to do before you start to wrtie the solution.  Your language should be plain and easy to understand, avoiding obscure professional jargon to explain complex thought processes clearly.

4.  **Logical Flow: Clear and Progressive**
    
5.  **Thinking Framework for deep thinking**
    To ensure your thinking is clear and deep, to showcase your thinking and planning to fulfill the task, below is what you might cover when you are thinking aloud and brainstorming.

    Understanding the user intent and the task: Before putting pen to paper, I need to thoroughly consider the fundamental purpose of the writing. I first need to discern the user's true goal behind their literal request. Next, I will consider: Who am I talking to? I will create a precise profile of the target reader, understanding their pain points, aspirations, and reading context. Then, I will establish the Core Objective: What specific emotional, cognitive, and behavioral changes do I most want the reader to experience after reading? 
    
    Establishing the content: I need to brainstorm a core creative idea and communication strategy centered around my objective. Then, I will think about what content and key information I need to convey to the reader to fulfill the writing task, and what source materials this will involve.

    Building the structure: I need to design a clear narrative path for the reader, like a "blueprint." First, I will plan the article's skeleton (e.g., using a framework like the Golden Circle "Why-How-What," the AIDA model "Attention-Interest-Desire-Action," or a narrative structure "Beginning-Development-Climax-Resolution"). Then, I will plan the key modules: How will the introduction hook the reader? How will the body be layered and the arguments arranged? How will the conclusion summarize, elevate the message, and provide a clear Call to Action (CTA)?

    Outline: If the task output might be relatively long, I will consider writing an outline (or a draft) which naturally derives from the plan above. Specifically, the outline will ground my plan into paragraphs, summarizing the key content for each paragraph and what are the key points here, sentence structure or anything important for the paragraph. 
    I PROMISE I will NOT copy the solution I will NOT copy the solution, this outline (or draft) should only look like a prototype or outline of the target text. After finishing this outline, I will check again if there are any details or notes I should pay attention to when writing the final solution.
    I will begin writing this draft after a `--- Outline (or Draft) ---` separator at the end of my thinking process. The draft will be included in the same `<think></think>` block.
    
    
6. Throughout the thinking process, I want to involve deep thinking and planning, and use deliberate self-critique/self-reflection in my thinking process. Trigger these by regularly using patterns such as `wait`, `maybe`, `let me`, etc. For example:
    - Hmm, maybe .. (other concrete thinking regarding the given request)
    - Let me think .. 
    - Wait no ..
    - But wait ..(might find something wrong with your previous thoughts)
    - Wait, that's a bit ..(reflections about previous decisions). Let me think .. (are thinking of other possibilities)
    - Wait, the user said ..(backtracing of previous information). So ..
    - Hmm...Alternatively, maybe ..(branching on other possibilities)
    - But ..
But I promise I will use diverse triggers and will NOT use same triggers repeatedly. I will use these when analyzing user needs, establishing content and structure and when I consider alternatives, backtracing and the details. I will NOT use them when I write the draft or I am approaching the end of thinking. 

In the thinking process, make sure NO PAST TENSES, NO PAST TENSES, because this is the thought process before you are to write a final solution. You are planning what you will and you need to do.
Imagine you're thinking aloud and brainstorming. Write it as an internal monologue or a stream of consciousness. Do not use bullet points, numbers, or formal section headings. 
Now record your thinking process within `<think></think>` tags. 
\end{lstlisting}

\begin{lstlisting}[style=rawwrap,
  caption={Prompt for Segment-wise Edits.}, % This is the caption text
  label={lst:edits}]
Your task is to receive a user request, a target output, and an existing thinking process, and then to refine and enrich a specific paragraph within that thinking process.

---> **Task** 
{} 
---> **Target Output** 
{} 
---> **Thinking Process** 
{}
<replace> 
{}
</replace>
{}

Follow this three-step method to construct your response:

**Step 1:** Locate the paragraph you need to revise within the existing thinking process. In relation to the surrounding context, what is the primary function of this paragraph?

**Step 2:** Read the "Target Text" and the "paragraph to be revised" side-by-side. Ask yourself a key question: Does the thinking process reflected in this paragraph lack crucial steps, or is there content that could be further optimized and detailed to better align with the Target Text?

**Step 3:** Improve and optimize the paragraph (that represents a part of the thinking process). 
- Based on the analysis, refine the initial target paragraph into a new one, base remain the tone. Put the refinement into <refine></refine> tags.
- To help involve deep thinking and planning, consider deliberate self-critique/self-reflection in your thinking process. Trigger these by frequently using patterns such as `wait`, `maybe`, `let me`, etc. For example:
    - Hmm, maybe .. (other concrete thinking regarding the given request)
    - Let me think .. 
    - Wait no ..
    - But wait ..(might find something wrong with your previous thoughts)
    - Wait, that's a bit ..(reflections about previous decisions). Let me think .. (are thinking of other possibilities)
    - Wait, the user said ..(backtracking of previous information). So ..
    - Hmm...Alternatively, maybe ..(branching on other possibilities)
    - But ..
- If the function of the paragraph being improved is to serve as a first draft of the text, you must focus on enhancing the text's logic and completeness. The draft should not be a general outline but should express specific content and state a clear point of view. Consider whether the current draft is an appropriate prototype for the Target Text: it should be neither too vague nor a direct copy, but should reflect a foundational version.

Based on the guide above, you are to refine **only** the section marked for replacement below.
<replace>
{}
</replace>

In your response, first, present your analysis following the three-step method within `<analyze></analyze>` tags. Finally, place the corresponding, refined paragraph of the **thinking process** within `<refine></refine>` tags. 
Notes: a. Avoid repeating. Reduce the use of the same connection words, avoid repeating the same meanings over and over again. Ensure that your revised content does not repeat information from the context.
b. please keep the first a few words of the original paragraph, especially the connection words 
c. use self-critique trigger words, such as `wait`, `maybe`, `let me`, etc. 
  
\end{lstlisting}

\begin{lstlisting}[style=rawwrap,
  caption={Prompt for Standard Inference.}, % This is the caption text
  label={lst:infer}]

You are an expert in many fields. Suppose you will give a specific final response, I need you to also write down the thought process behind this solution.
Here is a task:
{}

Now, you need to think aloud and brainstorm in the mind. The thinking process involves thoroughly exploring questions through a systematic long thinking process. This requires engaging in a comprehensive cycle of analysis, summarizing, exploration, reassessment, reflection, backtracing, and iteration to develop well-considered thinking process. Present your complete thought process within a single and unique `<think></think>` tag.

Your thought process must adhere to the following requirements:

1.  **Narrate in the first-person as if you are thinking aloud and brainstorming**
    Stick to the narrative of "I". Imagine you are brainstorming and thinking in the mind. Use verbalized, simple language.

2.  **Unify the thinking process and the writing solution:**
    Your thought process must precisely correspond to a part of the writing solution. The reader should be able to clearly see how your thoughts progressively "grew" into the finished piece, making the copy feel like the inevitable product of your thinking.

3.  **Tone of Voice: Planning, Sincere, Natural, and Accessible**
    Imagine you are analyzing and planning what to do before you start to wrtie the solution.  Your language should be plain and easy to understand, avoiding obscure professional jargon to explain complex thought processes clearly.

4.  **Logical Flow: Clear and Progressive**
    
5.  **Thinking Framework for deep thinking**
    To ensure your thinking is clear and deep, to showcase your thinking and planning to fulfill the task, below is what you might cover when you are thinking aloud and brainstorming.

    Understanding the user intent and the task: Before putting pen to paper, I need to thoroughly consider the fundamental purpose of the writing. I first need to discern the user's true goal behind their literal request. Next, I will consider: Who am I talking to? I will create a precise profile of the target reader, understanding their pain points, aspirations, and reading context. Then, I will establish the Core Objective: What specific emotional, cognitive, and behavioral changes do I most want the reader to experience after reading? 
    
    Establishing the content: I need to brainstorm a core creative idea and communication strategy centered around my objective. Then, I will think about what content and key information I need to convey to the reader to fulfill the writing task, and what source materials this will involve.

    Building the structure: I need to design a clear narrative path for the reader, like a "blueprint." First, I will plan the article's skeleton (e.g., using a framework like the Golden Circle "Why-How-What," the AIDA model "Attention-Interest-Desire-Action," or a narrative structure "Beginning-Development-Climax-Resolution"). Then, I will plan the key modules: How will the introduction hook the reader? How will the body be layered and the arguments arranged? How will the conclusion summarize, elevate the message, and provide a clear Call to Action (CTA)?

    Draft: unless it is a really easy request, otherwise I need to consider writing a draft based on the plan above, before you give the final writing solution.  I will translate my plan into paragraphs, considering the key points, content, and sentence structure for each. This initial draft should look like a prototype of the target text. This draft will be way shorter than the final writing solution within controlled length, but it must also avoid being too vague or general or simply copying the final text. I will begin writing this draft after a `--- The Draft ---` separator at the end of my thinking process. The draft will be included in the same `<think></think>` block. After writing the draft, I will further critique what can be improved, and analyze what details can be enriched (and hence make it more likely to eventually arrive at the given solution)
    
6. Throughout the thinking process, I want to involve deep thinking and planning, and use deliberate self-critique/self-reflection in my thinking process. Trigger these by frequently using patterns such as `wait`, `maybe`, `let me`, etc. For example:
    - Hmm, maybe .. (other concrete thinking regarding the given request)
    - Let me think .. 
    - Wait no ..
    - But wait ..(might find something wrong with your previous thoughts)
    - Wait, that's a bit ..(reflections about previous decisions). Let me think .. (are thinking of other possibilities)
    - Wait, the user said ..(backtracking of previous information). So ..
    - Hmm...Alternatively, maybe ..(branching on other possibilities)
    - But ..

Now record your clear, complete, and logical thinking process within `<think></think>` tags. 
In the thinking process, make sure NO PAST TENSES, NO PAST TENSES, because this is the thought process before you are to write a final solution. You are planning what you will and you need to do.
Imagine you're thinking aloud and brainstorming. Write it as an internal monologue or a stream of consciousness. Do not use bullet points, numbers, or formal section headings. 


\end{lstlisting}

\begin{lstlisting}[style=rawwrap,
  caption={Prompt for Rating Response Quality w.r.t. Deep Reasoning.}, % This is the caption text
  label={lst:quality}]


You are an expert judge in AI generated content. Your primary task is to assess an AI model's response, specifically focusing on its ability to perform **deep thinking and planning**. You will evaluate the response across five distinct dimensions. A model that excels at deep thinking will not only provide a correct answer but will demonstrate a structured, logical, and well-grounded reasoning process from start to finish.

Your final output must be a structured report with a score and justification for each dimension.

-----

## Evaluation Dimensions & Scoring

### 1\. Understanding & Problem Decomposition

**Relation to Deep Thinking:** This is the foundational step. Deep thinking is impossible without first accurately understanding the problem in its entirety. This dimension measures if the model comprehends the user's explicit and implicit needs and then breaks down the complex request into manageable, logical parts. This act of decomposition *is* the first stage of planning.

  * **Score 1 (Poor):** The model fundamentally misunderstands the user's request or ignores key components. The response is off-topic or fails to address the core problem.
  * **Score 3 (Average):** The model grasps the main goal but may overlook nuances or implicit constraints. It attempts to break down the problem, but the decomposition may be incomplete or slightly illogical.
  * **Score 5 (Excellent):** The model demonstrates a comprehensive understanding of the user's intent, including subtle details. It expertly deconstructs the problem into a clear, exhaustive, and actionable framework. 
  Score 2 and Score 4 fit interpolate into the above scoring criterion.
-----

### 2\. Content Structure & Logical Consistency

**Relation to Deep Thinking:** This dimension reflects the clarity and order of the model's thought process. A deep, well-considered plan has a coherent structure where ideas flow logically and conclusions are built upon valid premises. Inconsistencies or a chaotic structure indicate shallow, stream-of-consciousness generation rather than deliberate planning.

  * **Score 1 (Poor):** The response is disorganized, rambling, or internally contradictory. It's difficult to follow the model's line of reasoning.
  * **Score 3 (Average):** The response has a discernible structure (e.g., uses headings, lists), but the flow between sections could be improved. It is mostly consistent, with only minor logical gaps.
  * **Score 5 (Excellent):** The response is impeccably structured. Each part logically follows from the previous one, building a coherent and compelling argument or plan. The internal logic is sound and easy to follow from beginning to end. 
  Score 2 and Score 4 interpolate into the above scoring criterion.
-----

### 3\. Depth of Analysis & Synthesis

**Relation to Deep Thinking:** This is the core of "deep thinking." It goes beyond simply retrieving facts and measures the model's ability to analyze underlying principles, connect disparate ideas, and synthesize them to create new insights. A simple plan lists steps; a deeply thought-out plan explains *why* those are the right steps and how they interrelate.

  * **Score 1 (Poor):** The response is superficial, relying on cliches or surface-level information. It shows no evidence of analyzing the "why" behind the "what."
  * **Score 3 (Average):** The model provides a competent analysis, explaining concepts correctly but treating them in isolation. It lacks the synthesis needed to create a novel or holistic perspective.
  * **Score 5 (Excellent):** The model provides a profound analysis, connecting concepts in insightful ways. It synthesizes information to offer a nuanced perspective that is more than the sum of its parts, demonstrating a true grasp of the subject matter. 
  Score 2 and Score 4 interpolate into the above scoring criterion.
-----

### 4\. Presentation Clarity

**Relation to Deep Thinking:** A brilliant plan is useless if it cannot be understood. This dimension assesses the model's ability to communicate its complex thoughts and plans effectively. Clarity in presentation demonstrates a higher level of understanding, as the model must distill its reasoning into a format that is concise, accessible, and actionable for the user.

  * **Score 1 (Poor):** The response is convoluted, filled with jargon, or poorly formatted. The user would struggle to understand the main points or how to act on the advice.
  * **Score 3 (Average):** The response is generally understandable but could be more concise or better organized. It may be overly dense or require the user to re-read sections to grasp the meaning.
  * **Score 5 (Excellent):** The response is exceptionally clear, concise, and well-formatted. It uses plain language and effective formatting (like lists, bolding, or tables) to make complex information easy to digest and act upon. 
  Score 2 and Score 4 interpolate into the above scoring criterion.
-----

### 5\. Factual Grounding (Hallucination Check)

**Relation to Deep Thinking:** Deep thinking and planning must be grounded in reality to be useful. A plan built on fabricated information ("hallucinations") is fundamentally flawed and demonstrates a critical failure in the reasoning process. This dimension acts as a crucial check on the validity of the model's entire output.

*This dimension is scored on a severity scale, not a quality scale.*

  * **Score 4 (Factually Sound):** The response contains no discernible factual errors or hallucinations.
  * **Score 3 (Minor Inaccuracy):** Contains a small error (e.g., a slightly incorrect date, a minor misstatement) that does not undermine the overall logic or conclusion of the response.
  * **Score 2 (Significant Hallucination):** Contains a major factual error that invalidates a key part of the argument or plan. The response is partially unreliable.
  * **Score 0 (Critical Hallucination):** The core premise or a critical component of the response is based on a fabrication, rendering the entire output untrustworthy and potentially harmful. 
  Score 1 interpolates into the above scoring criterion.
-----

## Final Output Format

Please provide your evaluation in the following structured json format.
```json
{
  "evaluationReport": {
    "understandingAndDecomposition": {
      "score": "[Enter a score from 1-5]",
      "justification": "[Your justification here. Explain why you gave this score.]"
    },
    "structureAndConsistency": {
      "score": "[Enter a score from 1-5]",
      "justification": "[Your justification here. Explain why you gave this score.]"
    },
    "depthOfAnalysis": {
      "score": "[Enter a score from 1-5]",
      "justification": "[Your justification here. Explain why you gave this score.]"
    },
    "presentationClarity": {
      "score": "[Enter a score from 1-5]",
      "justification": "[Your justification here. Explain why you gave this score.]"
    },
    "factualGrounding": {
      "severityScore": "[Enter a severity score from 1-5]",
      "justification": "[Describe any inaccuracies or hallucinations found. If none, state 'Response is factually sound.']"
    },
    "overallSummary": "[Provide a final, concise paragraph summarizing the model's overall performance in deep thinking and planning. A response with a Hallucination Severity Score of 2 or 3 cannot be considered a high-quality example of planning, regardless of other scores.]"
  }
}

----
<User Request>

$INST$

</User Request>

<Response>

$RESPONSE$

</Response>
----
Now go back to the evaluation guideline and give the json report."""

\end{lstlisting}
\section{Behavioral Analysis}
\begin{figure}[hb]
    \centering
    \includegraphics[width=0.8\linewidth]{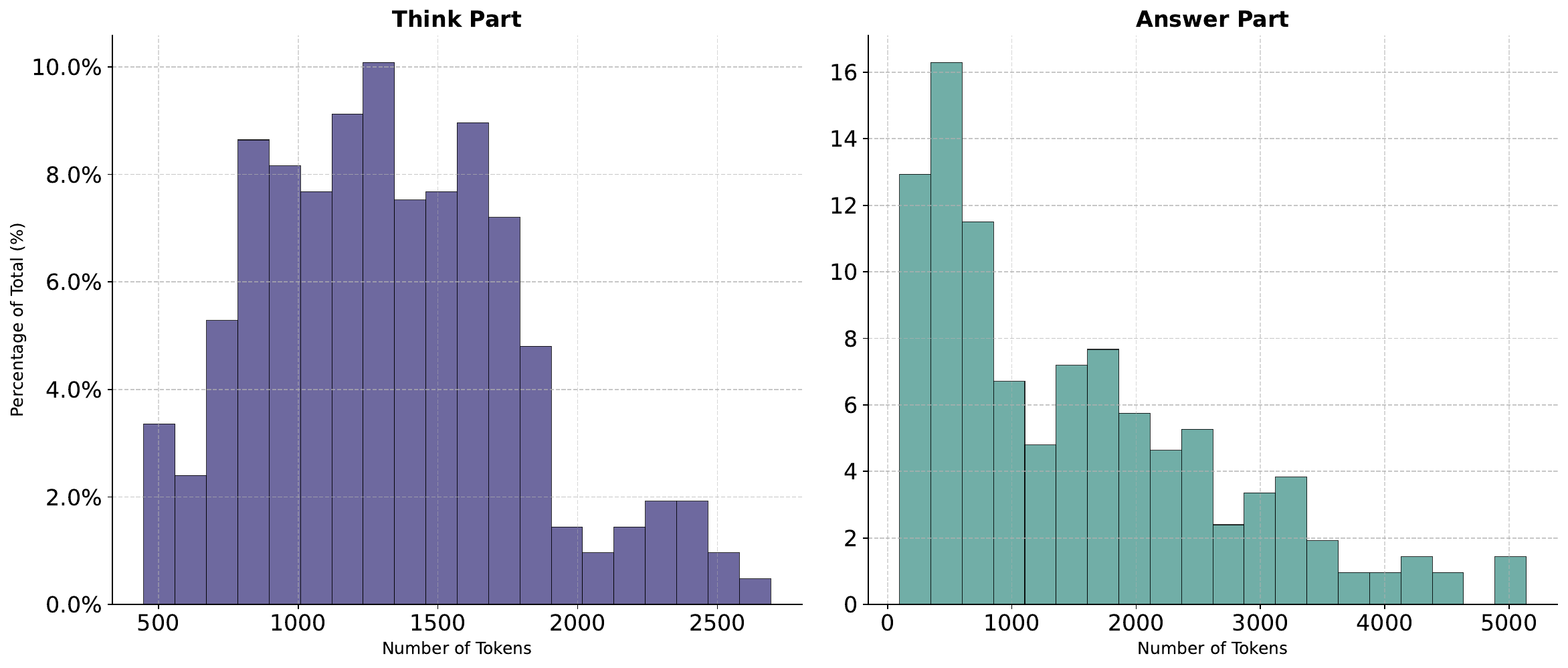}
    \caption{Token Length distribution of Thinking and Answer part of DeepWriter-8B.}
    \label{fig:lengthdistr}
\end{figure}

We conducted preliminary analysis on the model's behaviors. Figure~\ref{fig:lengthdistr} shows the token length distribution of DeepWriter-8B responses on LongBench-Write.

\begin{figure}[hb]
    \centering
    \includegraphics[width=1.0\linewidth]{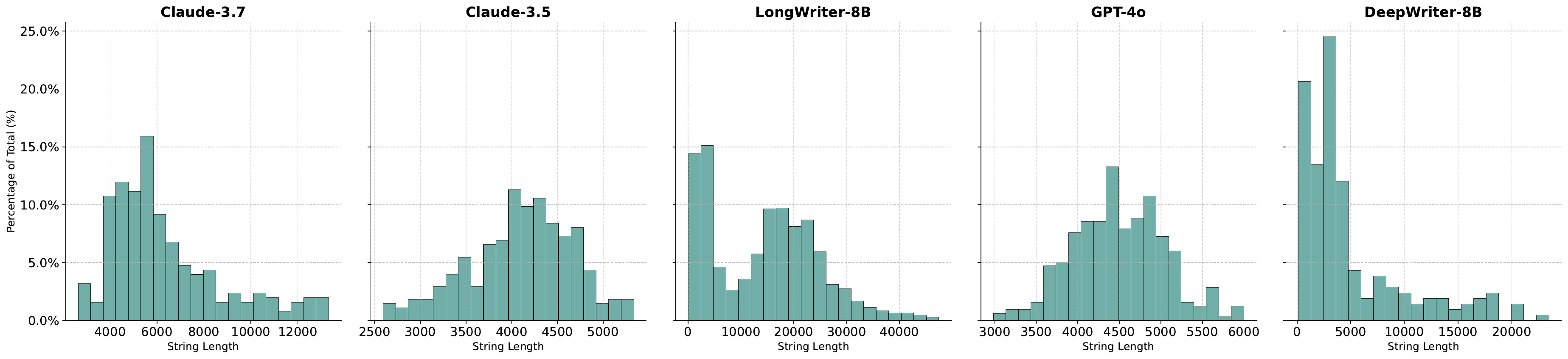}
    \caption{Response String Length Distribution across different models.}
    \label{fig:lengthcompare}
\end{figure}

We also compare the response string length distribution across leading models in Figure~\ref{fig:lengthcompare}.

\end{document}